\documentclass[fullpaper,final]{nldl}

\paperID{38}

\vol{265}

\usepackage{mathtools}

\usepackage{graphicx}

\usepackage{booktabs}

\usepackage{enumitem}

\usepackage{algorithm}
\usepackage[noend]{algpseudocode}

\usepackage{listings}
\usepackage{hyperref}
\usepackage{url}
\hypersetup{
  pdfusetitle,
  colorlinks,
  linkcolor = BrickRed,
  citecolor = NavyBlue,
  urlcolor  = Magenta!80!black,
}

\title{World Model Agents with Change-Based Intrinsic Motivation}
\author[1]{Jeremias Ferrao}
\author[1]{Rafael Fernandes Cunha}
\affil[1]{Faculty of Science, University of Groningen}
\affil[ ]{\texttt{j.l.ferrao@student.rug.nl}}

\begin{document}
\maketitle

\begin{abstract}
Sparse reward environments pose a significant challenge for reinforcement learning due to the scarcity of feedback. Intrinsic motivation and transfer learning have emerged as promising strategies to address this issue. Change Based Exploration Transfer (CBET), a technique that combines these two approaches for model-free algorithms, has shown potential in addressing sparse feedback but its effectiveness with modern algorithms remains understudied. This paper provides an adaptation of CBET for world model algorithms like DreamerV3 and compares the performance of DreamerV3 and IMPALA agents, both with and without CBET, in the sparse reward environments of Crafter and Minigrid. Our tabula rasa results highlight the possibility of CBET improving DreamerV3's returns in Crafter but the algorithm attains a suboptimal policy in Minigrid with CBET further reducing returns. In the same vein, our transfer learning experiments show that pre-training DreamerV3 with intrinsic rewards does not immediately lead to a policy that maximizes extrinsic rewards in Minigrid. Overall, our results suggest that CBET provides a positive impact on DreamerV3 in more complex environments like Crafter but may be detrimental in environments like Minigrid. In the latter case, the behaviours promoted by CBET in DreamerV3 may not align with the task objectives of the environment, leading to reduced returns and suboptimal policies.
\end{abstract}

\section{Introduction}

In Reinforcement Learning (RL), an agent's ability to efficiently identify and optimize reward-generating behaviors is crucial. Yet, conventional RL algorithms often struggle in sparse reward environments \citep{rl-book}. The lack of frequent feedback in these settings complicates learning, leading to extended exploration phases and slow convergence towards optimal solutions. This topic has become a significant focus in recent years, driving the development of increasingly sophisticated exploration strategies.

One promising approach to address the sparse reward problem is to incorporate intrinsic rewards, which are rewards generated by the agent itself, rather than the environment. Intrinsic rewards encourage exploration and learning, even in the absence of extrinsic rewards. For example, count-based methods \citep{count,hashCount} provide intrinsic rewards based on the novelty of states visited by the agent, view-based methods \citep{viewX} reward the agent for visiting unexplored regions of the environment based on an internal map, and curiosity-driven methods \citep{pathak2017curiositydriven, rnd,raileanu2020ride} reward the agent for making surprising predictions about the environment and rewards obtained. 

Transfer learning offers a complementary solution to the problem of sparse rewards \citep{rl-transfer,deep-transfer}. By reusing skills or knowledge acquired previously, an agent can potentially accelerate the exploration process in a new but related setting. This allows agents to exploit commonalities between environments, minimizing the need for exhaustive exploration from scratch (tabula rasa).

CBET (Changed Based Exploration Transfer) \citep{cbet} exemplifies a new evaluation paradigm that marries both intrinsic rewards and transfer learning. In CBET, a pre-trained model first explores an environment to identify interesting interactions, which is then used to guide a task-specific model to optimize extrinsic rewards. This approach leverages knowledge gained in prior environments to guide exploration more effectively, while also incorporating intrinsic rewards to encourage exploration and learning. The authors demonstrated encouraging results across various sparse reward environments, highlighting the potential of this approach in RL.

However, CBET has been evaluated primarily with the IMPALA algorithm. \citep{espeholt2018impala}. Since its publication, significant advancements have been made in RL, resulting in more sophisticated algorithms. DreamerV3 \citep{dreamerv3}, in particular, is of great interest in the field of RL due to its feat of being the first algorithm to obtain a diamond from scratch in Minecraft, a challenging sparse reward environment. The algorithm has achieved state-of-the-art performance in several environments with minimal hyper-parameter tuning required. While CBET was proposed for use in model-free algorithms, it remains unexplored in the context of world model agents like DreamerV3 making the algorithm a prime candidate for evaluation with CBET.


In our work, we provide an adaptation of the CBET framework to accommodate world model agents like DreamerV3 during transfer learning. Our experiments compare the obtained returns of DreamerV3 and IMPALA agents, with and without CBET, in both tabula rasa and transfer learning settings. We use the environments of Crafter \citep{hafner2021crafter} and Minigrid \citep{MinigridMiniworld23} for conducting our experiments. Our results suggest that CBET provides a positive impact on DreamerV3 in the crafter environment but is detrimental in Minigrid. Additionally, DreamerV3 converges to a suboptimal policy in Minigrid compared to IMPALA, highlighting that the environment's characteristics and model architecture significantly impact the effectiveness of exploration strategies. To facilitate replicability of our results, we make our code available at \url{https://github.com/Jazhyc/world-model-policy-transfer}.

\section{Methods}\label{sec:style}

This section outlines the key algorithms and methodologies used in our study. We provide a brief overview of the IMPALA and DreamerV3 algorithms, followed by our proposed extension of the CBET framework to evaluate policy transfer in sparse reward environments.

\subsection{IMPALA}

IMPALA is a distributed, model-free reinforcement learning algorithm designed to efficiently scale across multiple environments and learners \citep{espeholt2018impala}. It employs a centralized learner that receives experience from multiple actors (agents) operating in parallel. The algorithm uses V-trace to address the differences in policies executed by the actors and the policy learned by the centralized learner. This approach allows IMPALA to achieve high throughput and effective learning in complex environments, albeit with a significant reliance on extensive environment interactions to learn robust policies.

\subsection{DreamerV3}

DreamerV3 represents a state-of-the-art model-based reinforcement learning algorithm that leverages a world model to simulate environment dynamics and predict future outcomes \citep{dreamerv3}. The core of DreamerV3 is its world model, which is trained to represent the latent state of the environment as a Partially Observable Markov Decision Process (POMDP) \citep{rssm}. The world model, consisting of a Recurrent State-Space Model (RSSM), allows DreamerV3 to generate imagined trajectories by predicting future latent states and rewards, thereby enabling the agent to learn policies without requiring substantial direct interaction with the environment. This simulation capability results in significantly higher sample efficiency compared to model-free approaches like IMPALA.

\subsection{CBET}
\label{subsec:cbet}

The original CBET framework introduced a policy transfer mechanism for model-free agents, leveraging intrinsic rewards to guide learning. This intrinsic reward is dependent on the `interestingness' of states visited by the agents taking into account information from the previous state. Formally, the intrinsic reward is defined as observed in Equation~\ref{eq:cbet}.

\begin{equation}
    \label{eq:cbet}
    r_i(s) = \frac{1}{n(s) + n(c)}
\end{equation}

In this equation, $r_i(s)$ is the intrinsic reward for state $s$, $n(s)$ is the visitation count of state $s$, and $n(c)$ is a count for the number of times a change has occurred in the environment. Change is defined as the difference between the current state and the previous state. 

The authors proposed two evaluation approaches for CBET: a tabula rasa evaluation, where the agent is trained from scratch in the task environment using both intrinsic and extrinsic rewards, and a transfer evaluation, where the agent is pre-trained in an exploration environment using intrinsic rewards before being fine-tuned in the task environment using extrinsic rewards. The transfer evaluation aims to leverage the agent's prior knowledge to accelerate learning in the task environment. Equation \ref{eq:cbet-transfer} outlines the policy transfer mechanism used in the CBET framework.

\begin{equation}
    \label{eq:cbet-transfer}
    \pi_{TASK}(s,a) = \sigma(
        f_i(s, a) + f_e(s, a)
        )
\end{equation}

Here, $s$ and $a$ refer to the state and action at a given time step. $f_i$ is the policy network that is trained in the exploration environment using only intrinsic rewards. This network remains fixed during the fine-tuning phase in the task environment. $f_e$ is the policy network that is trained in the task environment using extrinsic rewards. During exploration, the policy is determined only by $f_i$. The policy $\pi_{TASK}(s,a)$ in the task environment is a combination of the outputs of both networks, with $\sigma$ denoting the softmax operator.

\subsection{Intrinsic Reward}

We adopt the intrinsic reward mechanism based on interestingness proposed by \citet{cbet} as formulated in Equation~\ref{eq:cbet} and utilize pseudocounts to estimate the novelty of states.  Similar to the original CBET paper, we hash the states as proposed by \citet{hashCount} to account for similar states in complex environments.

Mirroring the CBET approach, we also randomly reset the counts with a probability of $p \leq 1 - \gamma_i$ at each time step where $\gamma_i$ is the discount factor of the intrinsic reward ($0 <\gamma_i < 1$). This reset prevents the intrinsic reward from vanishing as the visitation count increases.  Importantly, these resets occur randomly during exploration rather than exclusively at episode boundaries. This prevents an artificial bias where initial states consistently yield higher intrinsic rewards compared to later states.

\subsection{Transfer Learning Architecture}\label{subsec:transfer-learning}

We employ a two-stage transfer learning process as outlined in Section~\ref{subsec:cbet}. For our IMPALA agent, we directly replicate the original CBET approach. However, we cannot apply the same approach to the world model-based DreamerV3 as CBET was designed for model-free algorithms (see Equation~\ref{eq:cbet-transfer}). In particular, the policy network and world model in DreamerV3 are tightly coupled. We would need to substantially modify the architecture for DreamerV3 to utilize two different policy networks which depend on a shared world model. To circumvent this limitation, we instead propose using two different instances of DreamerV3 and then averaging the outputs of their policy networks as seen in Equation~\ref{eq:world-transfer}. This approach allows us to maintain the original architecture of DreamerV3 while still incorporating the CBET framework.

\begin{align}
    \label{eq:world-transfer}
    \pi_{TASK}(x,a) = \sigma(
        f_i(w_i(x), a) + f_e(w_e(x), a)
        )
\end{align}

The primary modification in Equation~\ref{eq:world-transfer} is the inclusion of the world model $w$ in the policy transfer process. The world model is used to generate the latent state $z$ from the observation $x$ and is passed to the policy networks $f_i$ and $f_e$ to determine the probabilities of the agent's actions.

This strategy leverages the world model's learned environmental dynamics to guide the policy transfer process. Crucially, our modification operates at a high level of abstraction, allowing its application across diverse model-based architectures. We expected that this modification would enhance the transfer learning process, resulting in superior performance within the task environment.

\subsection{Environments}\label{subsec:environments}

To test our agents, we evaluate their performance in the Minigrid \citep{MinigridMiniworld23} and Crafter \citep{hafner2021crafter} environments. These environments are known for their sparse reward structures, making them ideal for assessing the efficacy of our tabula rasa and transfer learning approaches.

\subsubsection{Minigrid Worlds}

The Minigrid suite offers a collection of procedurally generated, grid-based environments designed to evaluate the generalization abilities of RL agents \citep{MinigridMiniworld23}. Agents operating in these environments must navigate rooms, gather keys, and unlock doors in an orthogonal maze to complete their goal. The environments are also stochastic with various elements in the environments being randomized between episodes such as the locations of walls and doors. The Minigrid World environments pose a significant challenge for traditional RL algorithms since rewards are only provided upon goal completion.  Furthermore, the agent has limited visibility with only the region in front of it visible necessitating a robust exploration strategy.

\begin{figure}[h]
    \centering
    \begin{subfigure}[b]{0.24\textwidth}
        \centering
        \includegraphics[width=\textwidth]{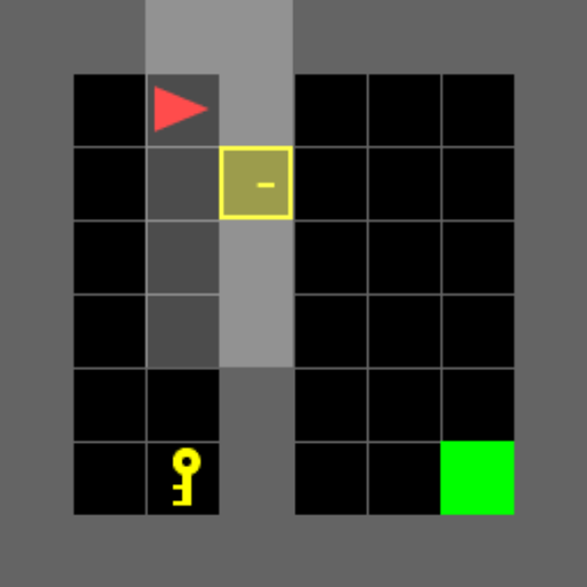}
        \label{fig:minigrid-doorkey}
    \end{subfigure}
    \hfill 
    \begin{subfigure}[b]{0.24\textwidth}
        \centering
        \includegraphics[width=\textwidth]{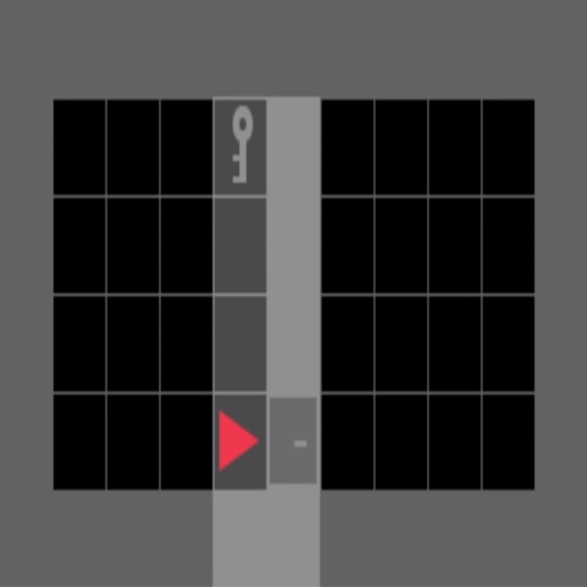}
        \label{fig:minigrid-unlock}
    \end{subfigure}
    \caption{Minigrid environments: Doorkey (left) and Unlock (right). The agent's observations (light coloured squares) consist of a 7 $\times$ 7 grid infront of it.}
    \label{fig:minigrid}
\end{figure}

Our main task environment is `Unlock'. Here, the agent must locate a key within a two-room layout and unlock a door. This environment is used for evaluating both the tabula rasa agent and the transfer agent after pre-training. The exploration environment is `Doorkey'.  This environment also contains a door and a key in the same room configuration. However, the objective is different and the agent must additionally reach the green square in the bottom-right of the maze.

\subsubsection{Crafter}

Crafter is a procedurally generated grid world that challenges agents to gather resources, craft items, and achieve specific goals \citep{hafner2021crafter}. To succeed in this environment, agents must be able to handle multi-step tasks effectively, as rewards are sparse and primarily granted upon completing achievements. There are 22 achievements in total, ranging in difficulty from crafting basic tools to attaining diamonds. The environment is characterized by partial observability, with the agent's view limited to a small grid centered on its position. Episodes in Crafter only end when the agent's health points reach zero, adding additional complexity to the agent's decision-making process.

\begin{figure}[h]
    \centering
    \includegraphics[width=0.4\textwidth]{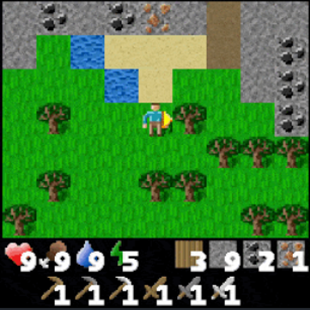}
    \caption{Crafter environment. The agent is provided a top down view of the game with statistics at the bottom.}
    \label{fig:crafter}
\end{figure}

A significant strength of the Crafter environment is its standardized budget of one million environment frames. This allows for direct performance comparisons across research papers, streamlining the evaluation of different agents. For our experiments, we pre-train transfer learning agents in a single environment (fixed seed) and subsequently assess their performance in environments with randomized seeds. We believe this approach sufficiently tests the agents' ability to generalize their acquired knowledge to novel configurations, providing a measure of their transfer learning capabilities.

\subsection{Experimental Setup}\label{subsec:experimental-setup}

We provide pseudocode for the tabula rasa and transfer learning approaches in Appendix \ref{subsec:pseudocode}. Our tabula rasa agents receive a computational budget of 3 million steps in Minigrid and 1 million steps in Crafter. We also perform 5 training runs for each combination of algorithm and environment resulting in 20 different experiments for the tabula rasa scenario. The rewards are averaged across these 5 training runs for evaluation.

Conversely, our transfer learning agents are pre-trained in the exploratory environment for one million steps before being fine-tuned in the task environment for an additional one million steps across both Minigrid and Crafter. Due to computational constraints, we were unable to conduct multiple runs for the transfer learning experiments. As a result, we only present the results of a single run for each algorithm and environment.

We perform minimal hyperparameter tuning for our agents, opting to use most of the default settings provided by the original CBET and DreamerV3 papers. Primarily, we needed to tune the intrinsic strength coefficient $\alpha$ for the tabula rasa agents as seen in Equation~\ref{eq:total-reward}. This coefficient controls the balance between intrinsic reward $r_i$ and regular extrinsic reward $r_e$, influencing the total reward $r_t$ and the agent's exploration behavior. At a coefficient of 0, there would be no difference between a regular RL algorithm and one modified with CBET. The appendix provides a detailed explanation of our hyperparameters and the results of our intrinsic reward scaling grid search.

\begin{equation}
    r_{t}(s) = r_{e}(s) + \alpha \cdot r_{i}(s)
    \label{eq:total-reward}
\end{equation}

\subsection{Evaluation Metrics}\label{subsec:evaluation-metrics}

For both Minigrid and Crafter environments, we evaluate our agents based on their cumulative extrinsic reward also known as the return. This metric allows us to determine the agent's overall performance in the environment, capturing its ability to complete tasks and maximize rewards. We utilize several evaluation episodes to assess the agent's performance over time as it progressively trains in the environment.

\section{Results}\label{sec:results}

\begin{figure*}
    \centering
    \includegraphics[width=0.8\textwidth]{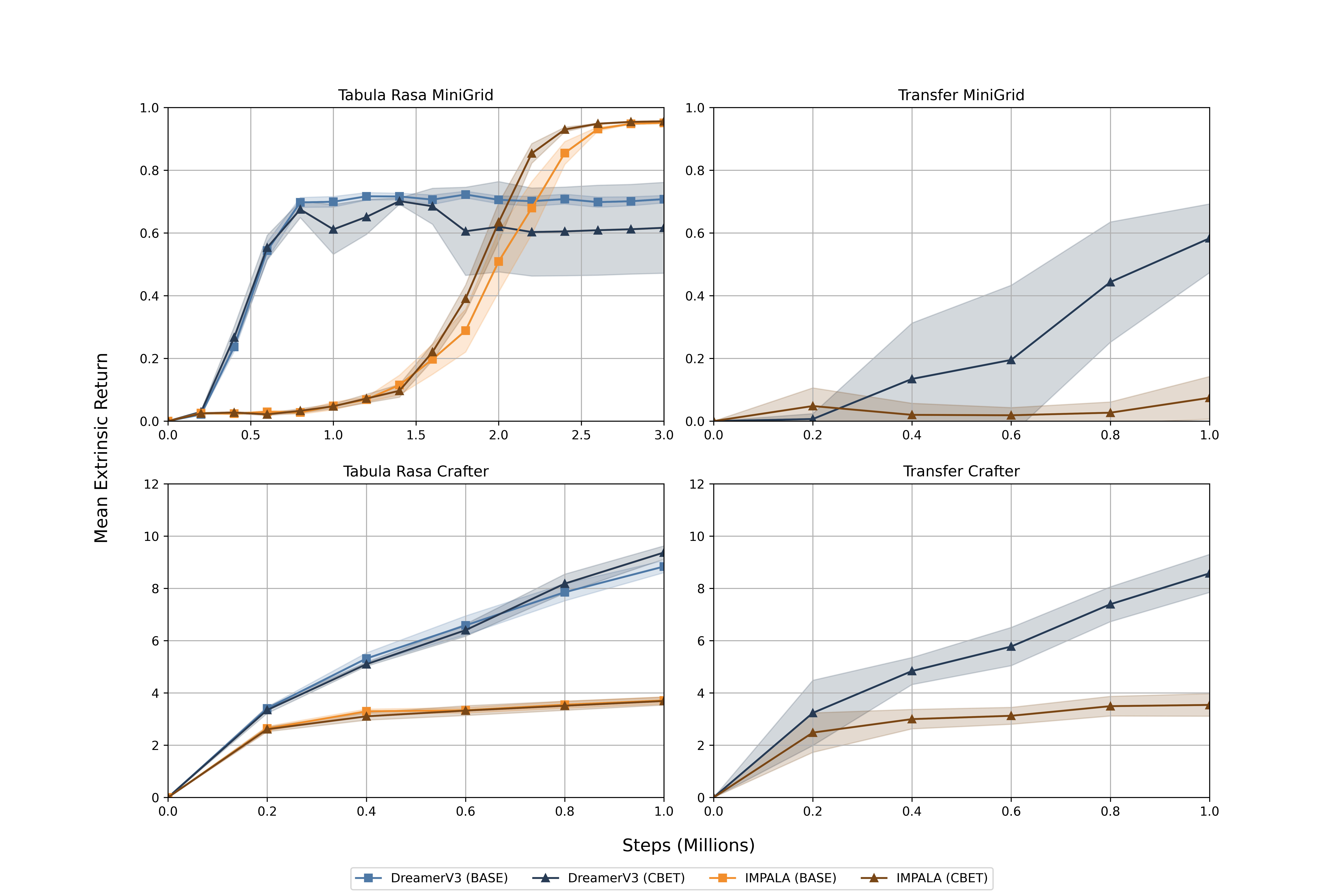}
    \caption{Mean extrinsic return plotted with standard error. Standard errors in the tabula rasa case represent variability across 5 experiments, while those in the transfer learning case reflect variability across 8 evaluation episodes. Transfer learning experiments exclusively feature the CBET variants of the algorithms. The transfer learning results indicate that DreamerV3 outperforms IMPALA in Crafter but IMPALA acquires higher returns initially in Minigrid. In the tabula rasa experiments, DreamerV3 outperforms IMPALA in Crafter but significantly falls short in Minigrid. CBET is also beneficial for DreamerV3 in Crafter but reduces returns and exhibits higher variance in Minigrid.}
    \label{fig:results}
\end{figure*}

Our results can be observed in Figure~\ref{fig:results} for the tabula rasa and transfer learning experiments. We use a rolling average of 200,000 steps to smooth the curves and provide a clearer representation of the agents' learning progress. The standard error is also included to provide an estimate of the variance in the results.

In the tabula rasa experiments, CBET produces a modest improvement in Crafter for DreamerV3 compared to the standard variant of the algorithm. However, the opposite effect is observed in Minigrid, where CBET reduces the agent's returns and substantially increases its variance. Interestingly, while DreamerV3 outperforms IMPALA in Crafter, the algorithm appears to converge to a suboptimal policy in Minigrid, with IMPALA surpassing DreamerV3 in terms of extrinsic returns.

For the transfer learning experiments, DreamerV3 attains near zero returns for the first 200,000 steps in Minigrid with IMPALA outperforming the algorithm during this time frame. However, DreamerV3 does eventually recover from this performance gap and surpasses IMPALA in terms of extrinsic returns. In Crafter, the same behaviour is not observed and DreamerV3 consistently outperforms IMPALA in the environment. 

\section{Discussion}\label{sec:contents}

The application of CBET to DreamerV3 is found to not be universally beneficial, proving advantageous in the more complex Crafter environment but detrimental in Minigrid. We arrived at this conclusion by observing the reduced returns in Minigrid during tabula rasa experiments and the near-zero returns during the initial stages of the transfer learning experiments. The latter suggests that pre-training DreamerV3 with the intrinsic CBET rewards does not directly lead to a policy that maximizes extrinsic rewards in Minigrid. When we combine these two results, it may be inferred that the behaviours promoted by CBET in DreamerV3 do not align with the task objectives of the Minigrid environment, leading to reduced returns. This behaviour is in contrast to the application of CBET in the Minigrid environment with IMPALA where the technique is beneficial. To our knowledge, CBET has only been evaluated with IMPALA and our work is the first to empirically determine its effectiveness with different RL algorithms. Thus, this behavioural finding is important as it suggests that the impact of CBET not only varies across different environments but also depends on the specific characteristics of the model architecture.

Additionally, DreamerV3 appears to have converged to a suboptimal policy in Minigrid, with IMPALA outperforming the algorithm in terms of extrinsic returns. This result may seem surprising at first given DreamerV3's state-of-the-art performance in most environments. However, even the original DreamerV3 paper demonstrates that the model occasionally attains lower returns compared to other algorithms in a narrow domain of environments. We hypothesize that IMPALA's large number of actors and asynchronous nature allows it to gather more diverse experiences and explore the environment more effectively.

While we provided an adaptation of CBET for world model algorithms like DreamerV3, this modification comes with a major caveat. DreamerV3 is a large RL algorithm requiring a significant amount of VRAM to train. Our adaptation of CBET doubles the already significant amount of VRAM required to train the model, as we create an additional world model and policy network. Future work could explore alternate transfer learning techniques like fractional transfer methods \citep{sasso2021fractionaltransferlearningdeep}, which involve selectively resetting and retraining components of the world model. This approach could potentially reduce the computational resources required for transfer learning in DreamerV3, making it more practical for a wider range of applications.

Moreover, both the standard and CBET variants of DreamerV3 underperformed compared to the results reported in the Crafter Experiments in the original DreamerV3 paper (reported return of 11.7 $\pm$ 1.9) \citep{hafner2021crafter}. This shortfall is likely due to the lower planning ratio used in our experiments, imposed by hardware limitations, which reduced the model's ability to simulate the environment and impacted the returns obtained.

The use of intrinsic rewards and the CBET mechanism, despite their benefits, presents significant challenges. Tuning the intrinsic strength coefficient is a delicate process — too high, and the agent overemphasizes exploration; too low, and exploration is insufficiently driven. Future research could focus on developing an intrinsic reward coefficient scheduler. Similar to learning rate schedulers, the agent begins with high intrinsic motivation to encourage exploration and gradually reduces it to promote exploitation. This approach, potentially combined with the ADAM optimizer \citep{Kingma2014AdamAM}, could reduce the need for extensive tuning. Further exploration and interpretation of the interaction between intrinsic rewards and transfer learning methods in more complex environments would also be valuable.

\section{Conclusion}\label{sec:conclusion}

Ultimately, our experiments reveal a nuanced picture of DreamerV3's performance and the impact of CBET. The application of the technique to DreamerV3 yields mixed results, proving beneficial in the more complex Crafter environment but detrimental in Minigrid. This variability in CBET's effectiveness across different environments and its interaction with DreamerV3 highlights the critical importance of carefully selecting and evaluating exploration strategies. Additionally, while DreamerV3 demonstrates superior performance in Crafter, it unexpectedly converges to a suboptimal policy in Minigrid compared to IMPALA, adding to a narrow set of environments where IMPALA outperforms DreamerV3. Our findings emphasize the need for a thoughtful approach when optimizing performance in sparse reward environments, as the impact of these techniques can vary significantly depending on the specific characteristics of the task and model at hand.

Addressing these issues could enhance the robustness of DreamerV3, enabling it to tackle a wider range of environments effectively. Developing a scheduler for intrinsic reward strength can optimize the exploration-exploitation balance over time, reducing the need for extensive parameter tuning. Moreover, efforts to reduce the computational resources required for DreamerV3 and modifications to the algorithm to utilize more actors which are trained asynchronously will make it more practical for a wider range of applications. Through these improvements, we can pave the way for more robust and versatile reinforcement learning models that can effectively tackle real-world problems.

\section*{Acknowledgments}

We thank the Center for Information Technology of the University of Groningen for their support and for providing access to the Hábrók high performance computing cluster.


\printbibliography

\appendix

\section{Computational Resources}\label{subsec:computational-resources}

All experiments mentioned in the main report were conducted for 1 million steps in Crafter and 3 million steps in Minigrid. Our DreamerV3 models were run on a single node with an Intel Xeon Gold 6150 CPU and Nvidia V100 GPU for approximately 12 and 48 hours in Minigrid and Crafter respectively. Our IMPALA models used a single node with 8 Intel Xeon Gold 6150 CPUs and an V100 GPU for approximately six hours in Crafter and 2 hours in Minigrid.

\section{Model and Experiment Hyper-parameters}\label{subsec:hyperparameters}

For DreamerV3, we used the implementation provided by the authors. Conversely, we used the TorchBeast implementation for IMPALA \citep{küttler2019torchbeastpytorchplatformdistributed}. We mainly utilize the same hyper-parameters as described in the original DreamerV3 and IMPALA papers \citep{dreamerv3,espeholt2018impala}. The number of actors which interact with the environment in IMPALA were reduced to 8 due to the limited amount of CPUs available to us. Likewise, we had to carefully select the planning ratio in DreamerV3 which determines the amount of time the model spends simulating the environment. We used a planning ratio of 64 along with the XL configuration of 200M parameters for DreamerV3. 

Both algorithms share the same evaluation strategy. 8 Evaluation episodes were scheduled every 10000 steps and the mean extrinsic return was recorded to obtain the results presented in the main report.

\section{Equivalent Training Time Comparison in Crafter}\label{subsec:training-time}

\begin{figure}[h]
    \centering
    \includegraphics[width=0.5\textwidth]{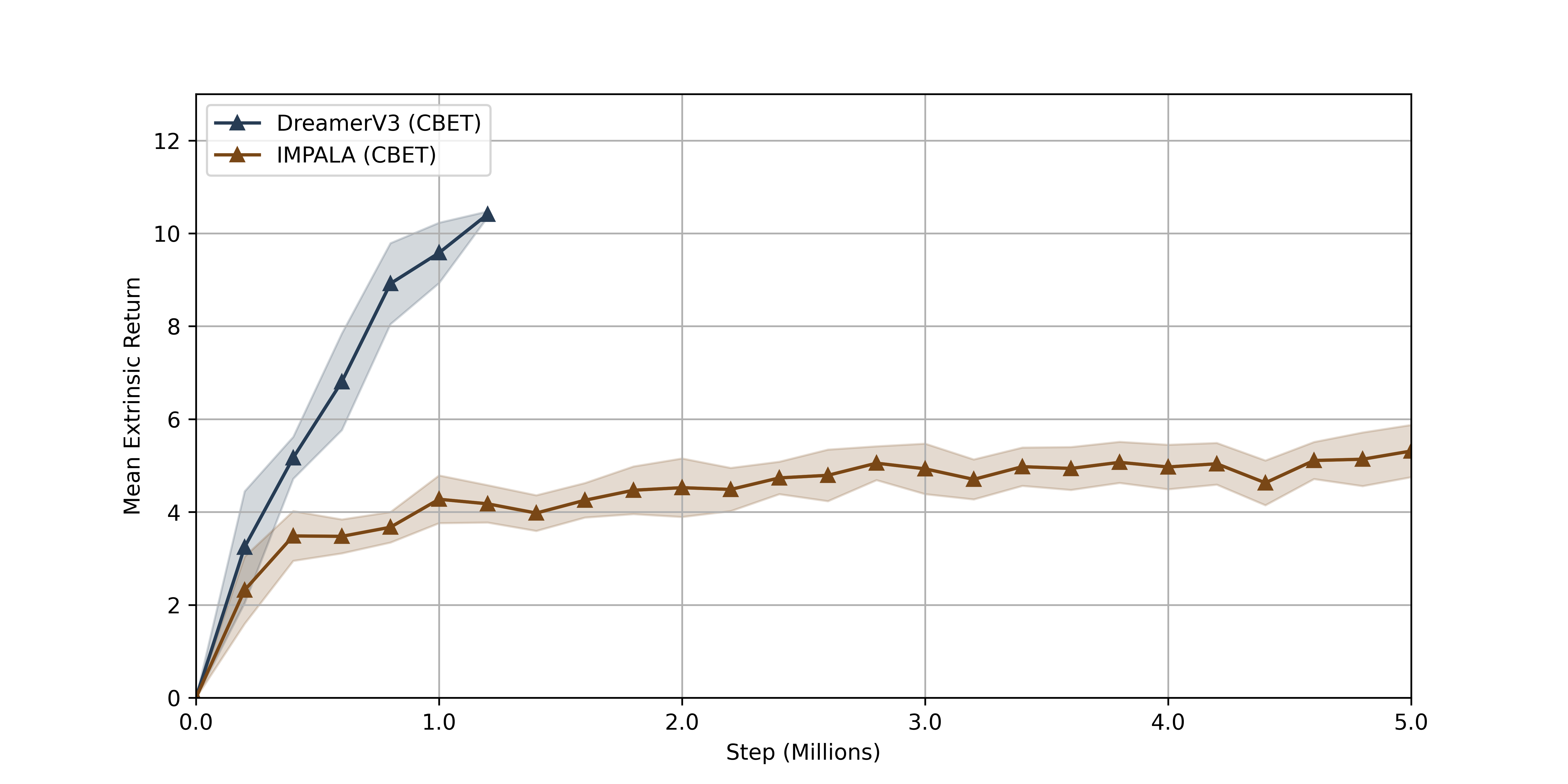}
    \caption{Equivalent Training Time Comparison between IMPALA and DreamerV3. IMPALA fails to outperform DreamerV3 even after being provided 5x more training time.}
    \label{fig:impala_vs_dreamerv3}
\end{figure}

We were curious how IMPALA would compare to DreamerV3 if we were to train them for an equivalent amount of time after observing the results in the Minigrid environment. We conduced an experiment in Crafter where both DreamerV3 and IMPALA were augmented with CBET and given a time budget of 15 hours. As seen in Figure~\ref{fig:impala_vs_dreamerv3}, DreamerV3 still manages to outperform IMPALA in terms of extrinsic reward. These findings attest to the high sample efficiency of DreamerV3. Nonetheless, it is important to note that IMPALA can be efficiently scaled to use more resources and potentially outperform DreamerV3 in the same time frame. The best choice of algorithm will depend on the available computational resources and the desired training time.

\section{Impact of Planning Ratio}\label{subsec:planning-ratio}

\begin{figure}
    \centering
    \includegraphics[width=0.5\textwidth]{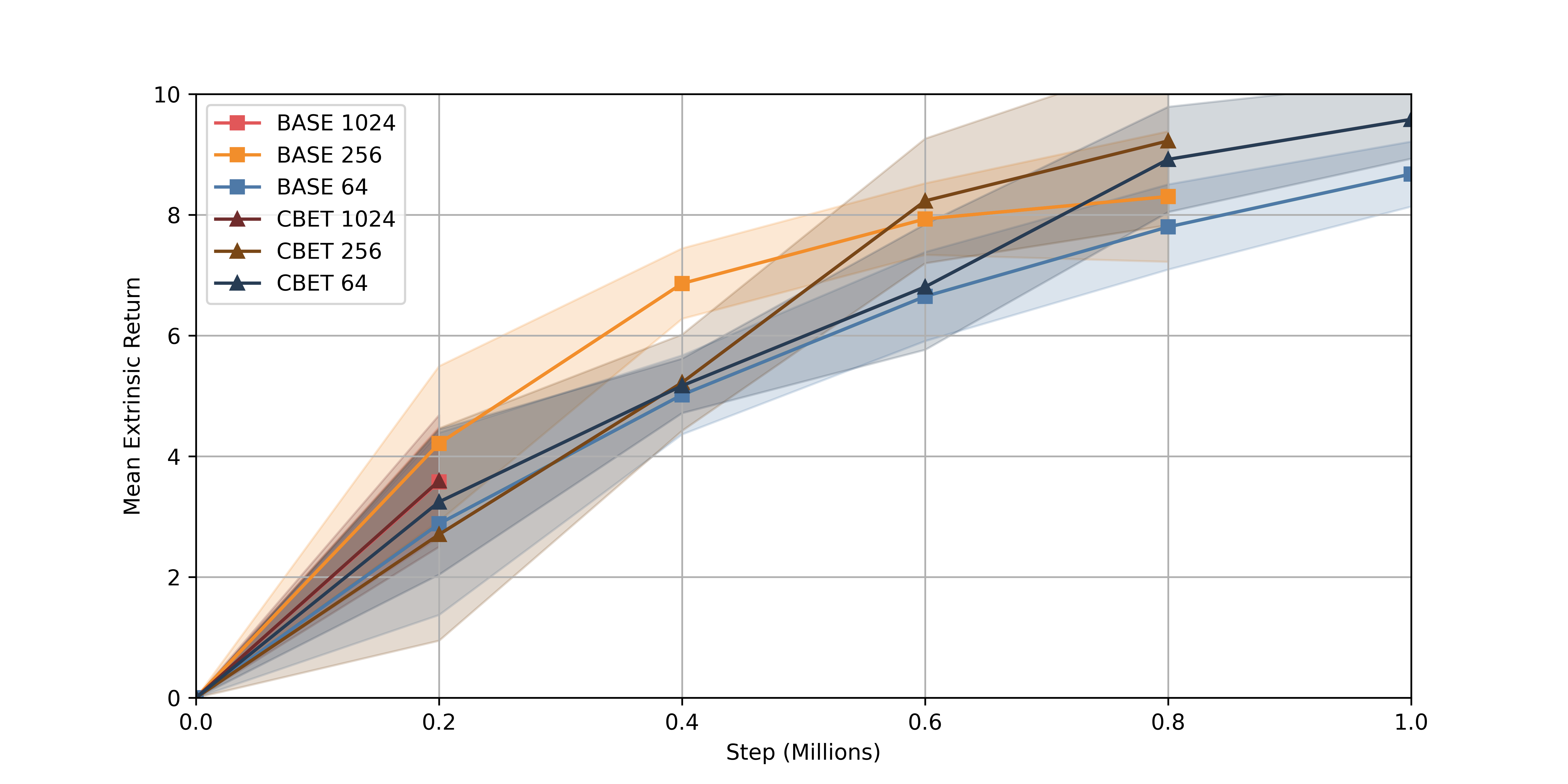}
    \caption{Impact of Planning Ratio on DreamerV3 with and without CBET in Crafter. There does not appear to be a significant difference in performance between the two models as the planning ratio increases.}
    \label{fig:planning-ratio-impact}
\end{figure}

We wished to determine the effect of the interaction between the planning ratio and the use of CBET on the performance of DreamerV3. We hypothesized that as the planning ratio increases, the gap between the performance of DreamerV3 with and without CBET would decrease. We arrived at this conclusion because a higher planning ratio allows the model to spend more time simulating and understanding the environment per step, potentially making the intrinsic rewards redundant.

To test our hypothesis, we performed a grid search in the Crafter environment using DreamerV3 with and without CBET. Only a single training run was used for each configuration. As the training time drastically increases with higher planning ratio, we applied a time limit of 24 hours for each configuration along with a limit of 1 million steps. The results of this experiment are shown in Figure~\ref{fig:planning-ratio-impact}. We observe that the gap between returns appears to decrease as the planning ratio increases. With a planning ratio of 1024, there is no substantial difference between the performance of DreamerV3 with and without CBET. However, we refrain from drawing definitive conclusions from this experiment due to our lack of data points and computational resources. We severely underestimated the real world time required during the planning phase of the model. As a result, the experiments were forced to terminate before reaching the desired number of steps.

\section{Intrinsic Reward Scaling}\label{subsec:intrinsic-reward-scaling}

\begin{figure*}
    \centering
    \includegraphics[width=0.9\textwidth]{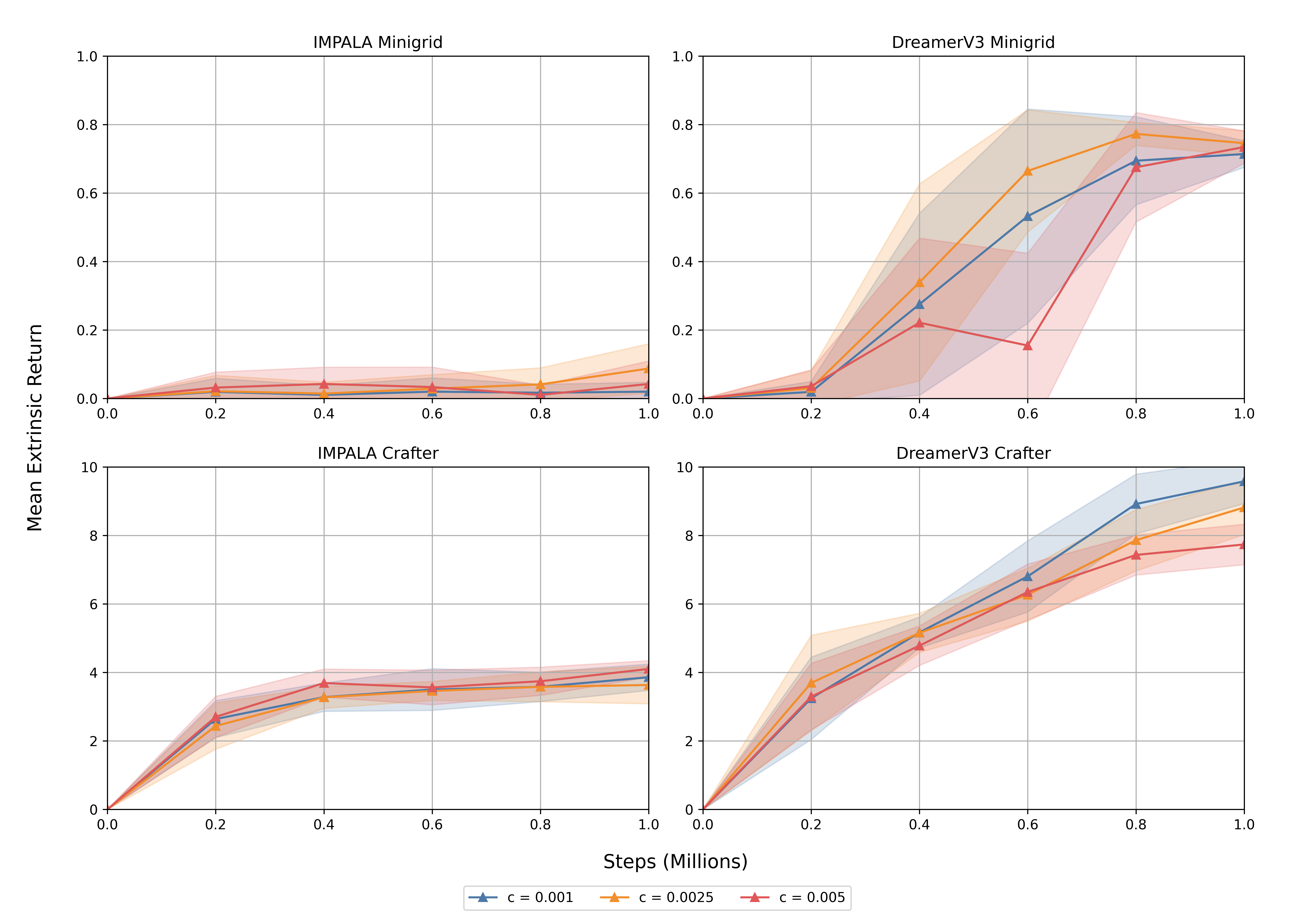}
    \caption{Results of the grid search for the intrinsic reward scaling factor $\alpha$. Extrinsic returns plotted with standard errors. Results summarized in Table~\ref{tab:comparison}.}
    \label{fig:intrinsic-reward-scaling}
\end{figure*}

To prevent the agent from focusing excessively on exploration at the expense of task completion, we needed to scale the intrinsic reward before adding it to the extrinsic reward. If this constant factor $\alpha$ is set too high, the agent will neglect its task. Conversely, if $\alpha$ is set too low, the intrinsic rewards may not sufficiently drive exploration. We performed a grid search using a single training run to find the optimal scaling factor among our candidates across both environments and algorithms. Figure~\ref{fig:intrinsic-reward-scaling} shows the results of this process. We selected the scaling factor that yielded the best performance in each case at the end of the training period. Our findings are summarized in Table~\ref{tab:comparison}.

\begin{table}[h]
    \centering
    \caption{Optimal intrinsic reward scaling factors among our candidates for each algorithm and environment from the grid search experiments.}
    \begin{tabular}{|l|l|c|}
        \hline
        \textbf{Method} & \textbf{Environment} & \textbf{Value} \\
        \hline
        IMPALA & Minigrid & 0.0025 \\
        DreamerV3 & Minigrid & 0.0025 \\
        IMPALA & Crafter & 0.005 \\
        DreamerV3 & Crafter & 0.001 \\
        \hline
    \end{tabular}
    \label{tab:comparison}
\end{table}

\section{Algorithm Pseudocode}\label{subsec:pseudocode}

In this section, we provide the pseudocode for the tabula rasa training and transfer learning algorithms used in our experiments. The tabula rasa algorithm is used to train the agents from scratch in the task environment, while the transfer learning algorithm is used to pre-train the agents in an exploration environment before fine-tuning them in the task environment. The algorithms are designed to work with both IMPALA and DreamerV3 agents and are adapted to incorporate the CBET framework.

\begin{algorithm}
\caption{Tabula Rasa Training with CBET}
\begin{algorithmic}[1]
\Require Agent, Environment
\State Initialize $agent$, environment, CBET counts
\Repeat
    \State Obtain action $a$ from current state $s$
    \State Execute action $a$, observe next state $s'$ and $r_e$
    \State Compute $c = s' - s$
    \State Compute $r_i(s) = \frac{1}{n(s) + n(c)}$ 
    \State Compute $r_t(s) = r_e(s) + \alpha \cdot r_i(s)$
    \State Update $agent$ using $s$, $a$, $s'$, $r_t$,
    \If{evaluation step is reached}
        \State Evaluate $agent$ using only $r_e$
    \EndIf
\Until{training is complete}
\end{algorithmic}
\end{algorithm}
    
\begin{algorithm}
\caption{Transfer Learning with CBET}
\begin{algorithmic}[1]
\Require Agent, Exploration environment, Task environment
\State Each agent has a policy network $f$. World model agents also have a world model $w$
\State Initialize $agent_i$, exploration environment, CBET counts
\State \textbf{Pre-training phase:}
\Repeat
    \State Obtain action $a$ from current state $s$
    \State Execute $a$, observe next state $s'$
    \State Compute $c = s' - s$
    \State Compute $r_i(s) = \frac{1}{n(s) + n(c)}$
    \State Update $agent_i$ using $s$, $a$, $s'$, $r_i$
\Until{pre-training is complete}
\State \textbf{Fine-tuning phase:}
\State Initialize $agent_{e}$ and task environment
\Repeat
    \State Observe $s$ in task environment
    \If{agent is world model-based}
        \State \scriptsize $\pi_{TASK}(x,a) = \sigma(f_i(w_i(x), a) + f_e(w_e(x), a))$ \normalsize
    \Else
        \State \scriptsize $\pi_{TASK}(s,a) = \sigma(f_i(s, a) + f_e(s, a))$ \normalsize
    \EndIf
    \State Execute $a$ from $\pi_{TASK}$, observe $s'$ and $r_e$
    \State Update $agent_{e}$ using $s$, $a$, $s'$, $r_e$
    \If{evaluation step is reached}
        \State Evaluate agent using only $r_e$
    \EndIf
\Until{fine-tuning is complete}
\end{algorithmic}
\end{algorithm}
    
\end{document}